\documentclass[twocolumn]{bmcart}
\usepackage{multirow}
\usepackage{graphicx}
\usepackage{amsthm,amsmath}

\usepackage[utf8]{inputenc} 

\startlocaldefs
\endlocaldefs

\begin{document}

\begin{frontmatter}
\begin{fmbox}

\dochead{Research}


\title{Extracting Lifestyle Factors for Alzheimer's Disease from Clinical Notes Using Deep Learning with Weak Supervision}


\author[
  addressref={aff1},                   
 noteref={n1},                        
  email={shenx499@umn.edu}   
]{\inits{z.s.}\fnm{Zitao} \snm{Shen}}
\author[
  addressref={aff1},
  noteref={n1},
  email={yixxx253@umn.edu}
]{\inits{Y.Y.}\fnm{Yoonkwon} \snm{Yi}}
\author[
  addressref={aff2},
  email={bompe001@umn.edu}
]{\inits{A.B.}\fnm{Anusha} \snm{Bompelli}}
\author[
  addressref={aff3},
  email={Fang.Yu.2@asu.edu}
]{\inits{F.Y.}\fnm{Fang} \snm{Yu}}
\author[
  addressref={aff4},
  email={Wang.Yanshan@mayo.edu}
]{\inits{Y.W.}\fnm{Yanshan} \snm{Wang}}
\author[
  addressref={aff2, aff5},
  corref={aff5},
  email={zhan1386@umn.edu}
]{\inits{R.Z.}\fnm{Rui} \snm{Zhang}}


\address[id=aff1]{
  \orgdiv{College of Science \& Engineering},             
  \orgname{University of Minnesota},          
  \city{Minneapolis},                              
  \cny{USA}                                    
}
\address[id=aff2]{%
  \orgdiv{Department of Pharmaceutical Care \& Health Systems},
  \orgname{University of Minnesota},          
  \city{Minneapolis},                              
  \cny{USA} 
}
\address[id=aff3]{%
  \orgdiv{Edson College of Nursing and Health Innovation},
  \orgname{Arizona State University},          
  \city{Phoenix},                              
  \cny{USA} 
}
\address[id=aff4]{%
  \orgdiv{Department of AI \& Informatics},
  \orgname{Mayo Clinic},          
  \city{Rochester},                              
  \cny{USA} 
}
\address[id=aff5]{%
  \orgdiv{Institute for Health Informatics},
  \orgname{University of Minnesota},          
  \city{Minneapolis},                              
  \cny{USA} 
}


\begin{artnotes}
\note[id=n1]{Equal contributor} 
\end{artnotes}



\begin{abstractbox}

\begin{abstract} 
\parttitle{Background} 
Since no effective therapies exist for Alzheimer's disease (AD), prevention has become more critical through lifestyle factor changes and interventions. Analyzing electronic health records (EHR) of patients with AD can help us better understand lifestyle's effect on AD. However, lifestyle information is typically stored in clinical narratives. Thus, the objective of the study was to demonstrate the feasibility of natural language processing (NLP) models to classify lifestyle factors (e.g., physical activity and excessive diet) from clinical texts.

\parttitle{Methods} 
We automatically generated labels for the training data by using a rule-based NLP algorithm. We conducted weak supervision for pre-trained Bidirectional Encoder Representations from Transformers (BERT) models on the weakly labeled training corpus. These models include the BERT base model, PubMedBERT(abstracts + full text), PubMedBERT(only abstracts), Unified Medical Language System (UMLS) BERT, Bio BERT, and Bio-clinical BERT.  We performed two case studies: physical activity and excessive diet, in order to validate the effectiveness of BERT models in classifying lifestyle factors for AD. These models were compared on the developed Gold Standard Corpus (GSC) on the two case studies.

\parttitle{Results} 
The PubmedBERT(Abs) model achieved the best performance for physical activity, with its precision, recall, and F-1 scores of 0.96, 0.96, and 0.96, respectively. Regarding classifying excessive diet, the Bio BERT model showed the highest performance with perfect precision, recall, and F-1 scores.

\parttitle{Conclusion} 
 The proposed approach leveraging weak supervision could significantly increase the sample size, which is required for training the deep learning models. The study also demonstrates the effectiveness of BERT models for extracting lifestyle factors for Alzheimer's disease from clinical notes.

\end{abstract}




\end{abstractbox}

\end{fmbox}

\end{frontmatter}


\section*{Background}
Alzheimer’s disease (AD) is the most common cause of dementia, accounting for 60 to 80 percent of all dementia cases \cite{association}. 
Around 5.8 million Americans were living with AD in 2020, and this number is expected to increase to approximately 14 million by 2050\cite{NIH}. Currently, no treatments can cure AD, but several lifestyle factors have been associated with a substantially reduced risk for AD with inconsistent findings. For example, high levels of physical and cognitive activity showed the strongest associations with reduced AD risk ranging from 11\% to 44\%\cite{frederiksen_gjerum_waldemar_hasselbalch_2019}. Multiple lifestyle modifications, including physical activity, no-smoking, light-to-moderate alcohol consumption, cognitive activities, and high-quality diets were correlated with a 60\% decreased risk for AD \cite{dhana2020healthy}. Furthermore, the Finnish Geriatric Intervention Study to Prevent Cognitive Impairment and Disability (FINGER) found that people at high risk of developing AD showed improvements in their cognitive abilities following two years of lifestyle changes \cite{kivipelto2013}. These findings led to the launch of multi-lifestyle intervention trials globally, such as the U.S. Pointer\cite{pointer}. Few Randomized Controlled Trials (RCTs) have the resources of the U.S. Pointer to be able to enroll a large sample. Hence, alternative, innovative, scalable, and cost-effective approaches to the use of electronic health records (EHRs) to establish causal effects are critically needed.  

Since 2009 when the Health Information Technology for Economic and Clinical Health Act (HITECH Act) was enacted\cite{HITECH}, EHRs have been adopted exponentially. Consequently, EHR studies have increased dramatically and have been acknowledged as a way of enhancing patient care and promoting clinical research \cite{WANG201834,VELUPILLAI201811,neveol,wu2017clinical}. EHRs document information obtained during healthcare delivery, including detailed explanations of the occurrence, treatment, and progression of diseases. Secondary analysis of observational EHR data has been widely used in multiple clinical domains \cite{critical2016secondary}.   

Much lifestyle information is documented in EHRs in the unstructured format, making it difficult to process and obtain desired information. To overcome this difficulty, natural language processing (NLP) techniques have been used to show promising results in extracting pertinent information from unstructured data for clinical research. For example, in our previous study, we demonstrated that extracting lifestyle factors using standard terminologies such as Unified Medical Language System (UMLS) and the existing NLP model (MetaMap) was feasible and reliable \cite{zhou_wang_sohn_therneau_liu_knopman_2019}. Our previous studies used standardized rule-based NLP models without the aid of annotated data. Besides, we demonstrated the feasibility of using NLP methods to automatically extract lifestyle factors from EHRs in our latest research, which was accepted by the HealthNLP 2020 workshop \cite{healthNLP}. Previously, conventional machine learning methods such as random forest, support vector machine (SVM), conditional random field, logistic regression, random forest (RF), bagged decision trees, and K-nearest Neighbors have been used for extracting lifestyle factors related to excessive diet, physical activity, sleep deprivation, and substance abuse. However, some limitations included working with a small-sized annotated corpus mainly due to the labor-intensive process of developing the corpus.

To reduce human efforts to generate annotations, weak supervision is one approach that trains machine learning models using weak labels generated by rule-based methods. Previously, Wang et al. \cite{wang2019clinical} have demonstrated the feasibility of using weak supervision and deep representation using word embeddings for clinical text classification tasks. Recently, more advanced neural network-based representations, such as Bidirectional Encoder Representation from Transformers (BERT)\cite{BERTORIGINAL}, have further improved NLP performance in multiple NLP downstream tasks, such as question answering. BERT models have been further applied in the biomedical domain, and its variances have been pre-trained on various biomedical corpora, such as biomedical literature and clinical records (e.g., MIMIC\cite{johnson2016mimic}), to gain a deep representation of biomedicine. In general, these domain-specific BERT models have shown promising performance on clinical applications \cite{pubmedbert,clinicalBERT,gu2020domain}.  For instance, Lee et al. \cite{DBLP:journals/corr/abs-1901-08746} presented that Bio BERT, pre-trained on large-scale biomedical corpora, significantly exceeds the standard BERT model on some popular biomedical NLP tasks, i.e., named entity recognition, relation extraction, and question answering. Michalopoulos et al. \cite{michalopoulos2020umlsbert} also showed similar results by comparing the general BERT model with their proposed domain-specific model, UMLS BERT. However, to the best of our knowledge, no clinical applications utilize BERT models with weak supervision so far. Hence, the objective of this study was to demonstrate the feasibility of using various pre-trained biomedical BERT models to classify lifestyle factors in clinical notes.  Our contributions include: 1) evaluating state-of-the-art clinical BERT models on the classification of lifestyle factors in clinical notes of patients with AD, and 2) using weak supervision to overcome the burdensome task of creating a hand-labeled dataset.

\section*{Methodology}
\begin{figure}[h!]
  \includegraphics[width=0.45\textwidth]{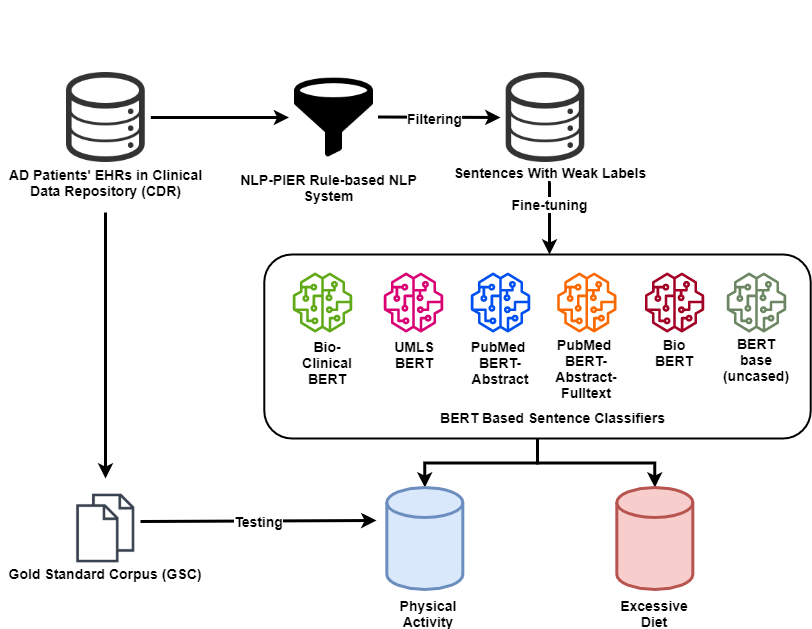}
  \caption{Overview of the study design}
  \label{fig:0}
\end{figure}

We conducted our experiments on two types of lifestyle factors: physical activity and excess diet. For each case study, we followed the same steps: 1) collecting clinical notes from patients with AD; 2) applying the rule-based NLP classifier to assign weak labels to the lifestyle-based sentences; 3) fine-tuning BERT models on the data with the weak label for each case study described below; 4) manually annotating a small portion of the selected sentences to develop the Gold Standard Corpus (GSC); 5) evaluating the performances of various BERT models in the GSC. Fig.\ref{fig:0} demonstrates the whole workflow for this study.

\subsection*{Data Source}
The clinical notes were sourced from the Clinical Data Repository (CDR) of the University of Minnesota (UMN). More than 180 million clinical notes are currently held by the aforementioned CDR, containing more than 2.9 million patients from 8 hospitals and more than 40 local clinics. Approval was obtained to access the EHRs for patients with AD from the Institutional Review Board (IRB).

For extracting sentences and generating weak labels for the sentences, we utilized NLP-PIER,  an information extraction (IE) platform to allow direct access to patient data stored in the free text of clinical notes in the CDR \cite{mcewan2016nlp}. The NLP-PIER provided direct access to patient data in the CDR as it used Elasticsearch technology and featured an open-source NLP system, BioMedICUS (BioMedical Information Collection and Understanding System)\cite{biomedicus3}. Regarding the rules that were used, we utilized a couple of lifestyle factors, such as physical inactivity and excessive diet, that had been found to be related to the development of AD. Then, by using the online UMLS Metathesarus browser based on our previous work, we manually collected all concept unique identifiers (CUIs) associated with physical activity and excess diet\cite{healthNLP}. These CUIs were used in the NLP-PIER to identify sentences with corresponding labels. 

 To generate the GSC, three annotators independently annotated 50 sentences for each case study using INCEpTION \cite{tubiblio106627}, a semantic annotation platform. Based on the presence of lifestyle factor entities of interest in the GSC,  the labels were assigned in the gold standard sentence level for this task. For physical activity, we set labels as "yes" or "no" ('no' indicates physical inactivity or no physical activity mentions) to sentences. For excess diet, sentences were assigned as "high calorie diet," "high salt diet," "high fat diet," or "normal/none" (indicating the normal diet or no diet mentions). The Cohen's Kappa scores reached 1.0 between any two annotators by the end. The entity names and their example can be found in Tab \ref{tab:0}. Note that while the GSC and training data with weak labels were mutually excluded by ensuring their note ids were not overlapped. 
\begin{table*}[t]
\caption{Example sentences with weak label for excessive diet and physical activity}
\resizebox{\textwidth}{!}{%
\begin{tabular}{|l|c|l|}
\hline
\multicolumn{1}{|c|}{\textbf{Category}}     & \textbf{Class}    & \multicolumn{1}{c|}{\textbf{Sentence Example}} \\ \hline
\multirow{4}{*}{\textbf{Excess Diet}} & High Fat Diet       & Pt is having fatty food ...                              \\ \cline{2-3} 
                                      & High Calorie Diet   & He had took high calorie diet for two weeks...           \\ \cline{2-3} 
                                      & High Salt Diet      & His current diet contains too much food with high salt.. \\ \cline{2-3} 
                                      & Normal /None        & She backs to normal diet...                              \\ \hline
\multirow{2}{*}{\textbf{Physical Activity}} & Physical Activity & Pt has increase regular physical activity...   \\ \cline{2-3} 
                                      & Physical Inactivity & He didn't maintain daily exercise...                     \\ \hline
\end{tabular}%
}
\label{tab:0}
\end{table*}

\subsection*{Model}

Transfer learning defines a process in which a model is trained on large size domain-related data set, then the model, as mentioned above's pre-trained weights, are fine-tuned on a small corpus for a specific task. Some benefits of transfer learning include how, for relatively high performance, less data is required. In this study, we evaluated six BERT models. The baseline model was the BERT base model, which was used as the baseline. Other five BERT models pre-trained in the domain-specific corpora include: PubMedBERT (pre-trained in the abstracts and full text of biomedical literature)\cite{gu2020domain}, PubMedBERT (pre-trained in only abstracts)\cite{gu2020domain}, Bio BERT\cite{DBLP:journals/corr/abs-1901-08746}, Unified Medical Language System (UMLS) BERT\cite{michalopoulos2020umlsbert}, and Bio-clinical BERT\cite{clinicalBERT}.  The use of the BERT model was a form of transfer learning. Besides the BERT base model, the other BERT models' other variations can be classified into two groups based on the pre-trained corpora. One group was pre-trained from scratch by using the text from biomedical literature in PubMed. More precisely, PubMedBERT(Abs+Ft) used the abstract and full text, which included approximately 16.8 billion words (107 GB). The PubMedBert(Abs) model was developed using the abstract alone, which included 3.2 billion words (21 GB). Similarly, Bio BERT was trained from PubMed abstracts and PubMed Central full-text articles. The remaining two variations of BERT models were trained on Medical Information Mart for Intensive Care III (MIMIC III) dataset \cite{johnson2016mimic}. The MIMIC III corpus has approximately 0.5 billion words (3.7 GB). Note that the Bio+Clinical BERT model was initialized from Bio BERT, and the UMLS BERT model was initialized from bio-clinical BERT while using the information from the CUIs.

\subsection*{Training and Evaluation}
For both case studies, we split the sentences into training(90\%), validation (10\%) in the weak-labeled corpus. During the testing, the random seed for all six models was kept the same. First, we fine-tuned BERT models on the training dataset and applied them to the validation set. The team fine-tuned all six BERT models on training data for ten epochs with a learning rate of $2*10^{-5}$. While keeping the trade-off on efficiency in mind, we picked a batch size of 512 for the case study on physical activity. The batch was set as 64 for the excessive diet case study. We used a dropout layer for regularization purposes and a fully connected layer in the end. The dropout rate was set to be 0.3, while the optimizer was set as Adam. The cross-entropy loss function was used as the loss function. In terms of padding, we padded the sentence to length 50 since the majority of sentences were shorter than that. The best models in the validation set were further evaluated in the GSC.


\section*{Results}

\subsection*{Corpus statistics}
We here describe the corpora for the two case studies to evaluate BERT models' effectiveness with weak supervision.

\subsubsection*{Case 1: Physical activity}
 We extracted 23,559 sentences by searching for physical activity related CUIs on the NLP-PIER. Of all the sentences, 22,785 (96.7\%) sentences were assigned as "physical activity," and 777 (3.3\%) mentions were "physical inactivity." In the GSC, the number of sentences in the category of physical activity and physical inactivity is 78 and 122, respectively.

\subsubsection*{Case 2: Excessive diet}
In total, 886 sentences were used as training data with weak labels. Training data were distributed as 302(34\%), 133(15\%), 153(17.3\%), and 300(33.7\%), respectively, for high calorie diet, high fat diet, high salt diet, and normal/none. In the GSC, the corresponding numbers were 18, 20, 20, and 70, respectively. 

\subsection*{Model performance}
During the experiment, the whole process was conducted on the Intel\textsuperscript{\tiny\textregistered} Xeon\textsuperscript{\tiny\textregistered}  Gold 6152 Processor with 22 cores and 256 RAM. In the first case study, it took about 12 and 18 hours to train one BERT model. For the excessive diet case study, this process took 0.5 to 2 hours, which was much shorter due to the relatively smaller training sample size compared to the first study. 

As shown in Tab \ref{tab:1}, all pre-trained clinical BERT models outperformed the BERT base model for physical activity.PubmedBERT (Abs) model performed the best with precision, recall, and macro F-1 score were 0.96, 0.96, and 0.96, respectively. The Bio-clinical BERT model metric was 9\%, 11\%, and 10\% higher, respectively, compared to the BERT Base model, which had precision, recall, and F-1 scores of 0.88, 0.86, and 0.87, respectively, and all other model surpassed the base model. 

\begin{table}[h!]
\caption{Comparison of results for various BERT models for the second case study (physical activity).}
  \begin{tabular}{cccc}
    \hline
    & Precision  &Recall   & F1\\ \hline
     BERT Base & 0.88 & 0.86 & 0.87\\
    PubmedBERT(Abs+Ft) & 0.92 & 0.90 & 0.91\\
    
     PubmedBERT(Abs) & \textbf{0.96} & \textbf{0.96} & \textbf{0.96} \\
    Bio BERT & 0.88 & 0.90 & 0.88\\
    
      UMLS BERT& 0.93 & 0.92 & 0.93\\
    Bio-clinical BERT& 0.89 & 0.88 & 0.89\\
    \hline
  \end{tabular}
  \label{tab:1}

\end{table}

\begin{table}[h!]
\caption{Comparison of results for various BERT models for the second case study (excessive diet).}
  \begin{tabular}{cccc}
    \hline
    & Precision  &Recall   & F1\\ \hline
     BERT Base & 0.94 & 0.97 & 0.95\\
    PubmedBERT(Abs+Ft) & 1.00 & 0.99 & 0.99\\
    
     PubmedBERT(Abs) & 0.85 & 0.90 & 0.83\\
    Bio BERT & \textbf{1.00} & \textbf{1.00} & \textbf{1.00}\\
    
      UMLS BERT& 0.79 & 0.86 & 0.77\\
    Bio-clinical BERT& 0.83 & 0.92 & 0.85\\ \hline
  \end{tabular}
  \label{tab:2}
\end{table}

In the use case on the excessive diet, the BERT base model performed very well with precision, recall, and F-1 scores of 0.94, 0.97, and 0.95, respectively Tab \ref{tab:2}. Only PubMedBERT and Bio BERT outperform the baseline model with minimal improvement. Bio BERT model reached excellent performance with all its precision, recall, and F-1 scores of 1.00. UMLS BERT performed the worst with significantly lower performance. The performance difference between BERT models was also wider, unlike the previous case study, which had a relatively similar result for all models.

\section*{Discussion}

The wide adoption of EHR in the healthcare system generates big data about the healthcare delivery of patients. The rapid advancement of artificial intelligence (AI) methods and computational resources provide an alternative way to cost-effectively consider the impact of lifestyle factor on AD using rich EHR data\cite{zhang2017advancing}. In this study, we demonstrated the feasibility of using BERT models with weak supervision to classify lifestyle factors for AD in clinical notes. The approach described in this study can be further extended to other lifestyle factors,  accelerating our investigation on the roles of lifestyle factors on AD.

To effectively train deep neural network models, extensive training data is required. Similar to other studies\cite{wang2019clinical}, weak supervision can generate sufficiently large training data by a rule-based NLP system without further human efforts. Although weakly labeled training data contain a certain noisy label, our finding demonstrated BERT models' promising performance on weakly labeled data on two classification tasks for lifestyle factors. Although these BERT models were pre-trained in the biomedical or clinical domain, their performances on different tasks are still distinguishable. For example, in the physical activity,  and PubMedBERT(Abs) outperformed PubMedBERT(Abs+Ft); however, the latter model performed better in the use case of classifying the excessive diet. In addition, the BERT model and its variations were robust to a dataset with imbalanced classes since all of them had impressive outcomes with a class ratio of 1:30.

During the evaluation of the models and error analysis, we found most BERT models' variations had difficulties in classifying the sentences in a complicated contextual situation. For example, the sentence "...states she stayed physically active with gardening and housework, but would like to increase her aerobic exercise..." meant that the patient was physically active. Still, she would like to be even more active. However, all models predicted the sentence mentioned above as "physical inactivity" while it should be "physical activity." Similarly, all models classified the sentence, "Patient continues to be physically active without doing any aerobic exercise outside of cardiac rehab" as "physical inactivity." At the same time, it should have been labeled as "physical activity." We believe that the phrase "without doing any aerobic exercise" may have led the model to mistakenly classify this sentence as it introduces a piece of contradicting information to the first part of the sentence. Besides, some sentences such as "The patient was very physically inactive," which seemed to be obvious to be "physical inactivity," were mispredicted as "physical activity." We believe that more extensive training dataset could avoid these errors.

Regarding the second case study (excessive diet), most of the sentences classified incorrectly are from the class of"normal/none." These were incorrectly classified as "high salt" and "high fat" for the most part. For example, the sentence, "history of cocaine abuse and acute syphilis" was classified as "high salt" when it should have been "none." It is possible that some domain specific BERT models didn't have enough knowledge on distinguishing the sentences with no mentions on life factors, comparing with the BERT Base model, which was trained on with broader domain corpus. In fact, for these specific sentences, a rule-based classification model would perform better. 

While our performance metric was relatively high (0.97-98) for both case studies, our study has some limitations. The first limitation of this study is that BERT models are sensitive to the training data size. The size of the excessive diet dataset, which was 886, was still relatively small. In addition to the differentiation within the model construction and pre-trained corpus, this may explain the presence of significant differences between BERT models and their variations than those in the first case study. Second, the highly imbalanced class in the case study on physical activity could be a limitation. Our results and limitations collectively indicate that our approach performed well, but further replication trials are needed to understand all models better.

Collectively, our findings have significant implications for further lifestyle research in AD. They provide a methodology to use unstructured EHR data to address the inconsistent findings of the strengths of association between lifestyle factors and AD risk and allow the simultaneous examinations of multiple lifestyle factors and their interactive/synergistic effects on cognitive changes and AD risk. Besides, they provide an approach for future casual modeling of lifestyle changes on clinical outcomes in AD. Since EHRs offer a potential source of data, it can be evaluated and defined to address questions that aim to measure the causal effect of intervention or exposure on the outcome of interest. Unlike RCTs that are time-consuming and expensive to carry out and have minimal generalizability, conducting studies using EHR data is scalable and affordable. Developing causal modeling methods using EHR data will allow large-scale and pragmatic trials.

\section*{Conclusion}
In this study, we used weak supervision and various pre-trained clinical BERT models to classify lifestyle factors in AD from clinical notes. The purpose of using weak supervision was to prevent the need for the laborious task of creating the hand-labeled dataset. We evaluated two text classification case studies' effectiveness: classifying sentences regarding their physical activity and excessive diet. We tested one baseline model and five different BERT models: PubmedBERT(Abs+Ft), PubmedBERT(Abs), Bio BERT, UMLS BERT, and Bio-clinical BERT. PubmedBERT(Abs) and Bio BERT model performed the best for the two use cases. The study group can further expand this approach to other factors such as substance abuse to investigate their effects on AD and provide additional AD research opportunities.

\begin{backmatter}
\section*{Funding}
Y.Y. thanks the University of Minnesota's Undergraduate Research Opportunities Program (UROP). This work was partially supported by the National Institutions of Health’s National Center for Complementary \& Integrative Health (NCCIH), the Office of Dietary Supplements (ODS) and National Institute on Aging (NIA) grant number R01AT009457 (PI: Zhang) and Clinical and Translational Science Award (CTSA) program grant number UL1TR002494 (PI: Blazar). The content is solely the responsibility of the authors and does not represent the official views of the NCCIH, ODS or NIA.

 \section*{Abbreviations}
 AD: Alzheimer’s Disease; BERT: Bidirectional Encoder Representations from Transformer; CUIs:Concept Unique Identifiers; GSC: Gold Standard Corpus; MIMIC III: Medical Information Mart for Intensive Care III; UMLS: Unified Medical Language System;



\section*{Competing interests}
There is no competing interests.





\bibliographystyle{bmc-mathphys} 
\bibliography{bmc_article}      

\end{backmatter}
\end{document}